\documentclass[lettersize,journal]{IEEEtran}
\usepackage{amsmath,amsfonts}
\usepackage{algorithmic}
\usepackage{algorithm}
\usepackage{array}
\usepackage[caption=false,font=normalsize,labelfont=sf,textfont=sf]{subfig}
\usepackage{textcomp}
\usepackage{stfloats}
\usepackage{url}
\usepackage{verbatim}
\usepackage{graphicx}
\usepackage{cite}
\usepackage{orcidlink}

\usepackage{amssymb}
\usepackage{mathtools}
\usepackage{amsthm}
\usepackage{eucal}
\usepackage{multirow}
\usepackage{adjustbox}
\usepackage{arydshln}
\usepackage{wrapfig}
\usepackage{float} 
\usepackage{subcaption}
\usepackage[capitalize,noabbrev]{cleveref}
\begin{document}

\title{When Normality Shifts: Risk-Aware Test-Time Adaptation for Unsupervised Tabular Anomaly Detection}

\author{Wei Huang~\orcidlink{0000-0003-2262-9339}, Hezhe Qiao~\orcidlink{0000-0003-3511-0528}, Kailai Zhang~\orcidlink{0009-0008-6067-2702}, Zaisheng Ye~\orcidlink{0000-0001-9881-6400}, Yu-Ming Shang\dag~\orcidlink{0000-0003-2903-4223}, Xiangling Fu\dag ~\orcidlink{0000-0002-1492-2829}
\thanks{\dag~Corresponding author.}

        % <-this % stops a space
\thanks{This paper was produced by the IEEE Publication Technology Group. They are in Piscataway, NJ.}% <-this % stops a space
\thanks{Manuscript received May 11, 2026.}
}
\markboth{Journal of \LaTeX\ Class Files,~Vol.~14, No.~8, August~2021}%
{Shell \MakeLowercase{\textit{et al.}}: A Sample Article Using IEEEtran.cls for IEEE Journals}


\maketitle

\begin{abstract}
Unsupervised tabular anomaly detection methods typically learn feature patterns from normal samples during training and subsequently identify samples that deviate from these patterns as anomalies during testing. However, in practical scenarios, the limited scale and diversity of training data often lead to an incomplete characterization of normal patterns. While test-time adaptation offers a remedy, its isolated focus on test-time optimization ignores the critical synergy with training-phase learning. Furthermore, indiscriminate adaptation to unlabeled test data inevitably triggers anomaly contamination, preventing the model from fully realizing its discriminative capability between normal and anomalous samples. To address these issues, we propose RTTAD, a \underline{R}isk-aware \underline{T}est-time adaptation method for unsupervised \underline{T}abular \underline{A}nomaly \underline{D}etection. RTTAD holistically tackles normality shifts via a synergistic two-stage mechanism. During training, collaborative dual-task learning captures multi-level representations to establish a robust normal prior. During testing, a Test-Time Contrastive Learning (TTCL) module explicitly accounts for adaptation risk by selectively updating the model using high-confidence pseudo-normal samples while constraining anomalous ones. Additionally, TTCL incorporates a k-nearest neighbor-based contrastive objective to refine embedding distributions, thereby further enhancing the model's discriminative capacity. Extensive experiments on 15 tabular datasets demonstrate that RTTAD achieves state-of-the-art overall detection performance.
\end{abstract}


\section{Introduction}
\IEEEPARstart{U}{nsupervised} tabular anomaly detection plays a pivotal role in ensuring the reliability of diverse real-world systems, such as medical diagnosis~\cite{fernando2021deep}, network intrusion detection~\cite{qiao2023truncated, qiao2024generative, niu2024zero}, financial fraud detection~\cite{al2021financial, qiao2024deep, qiao2025anomalygfm}, and industrial inspection~\cite{liu2024deep}. 
These unsupervised methods operate by characterizing the intrinsic feature patterns of normal samples during the training phase, and subsequently identifying instances that deviate from the established representation space as anomalies during testing.

Existing unsupervised anomaly detection methods for tabular data can be broadly classified into four categories: one-class classification methods~\cite{scholkopf1999support,ruff2018deep}, clustering/feature-distribution-based methods~\cite{liu2022unsupervised,ali2024anomaly,li2022ecod}, reconstruction-based methods~\cite{schlegl2017unsupervised,gong2019memorizing}, and self-supervised learning methods~\cite{shenkar2022anomaly,yin2024mcm,ye2025drl}. Despite the diversity and sophistication of these design strategies, they universally rely on a rigid fundamental assumption: the normal distribution observed during testing remains strictly consistent with that encountered during training.
\begin{figure}[t]
    \centering
    \includegraphics[width=0.9\columnwidth]{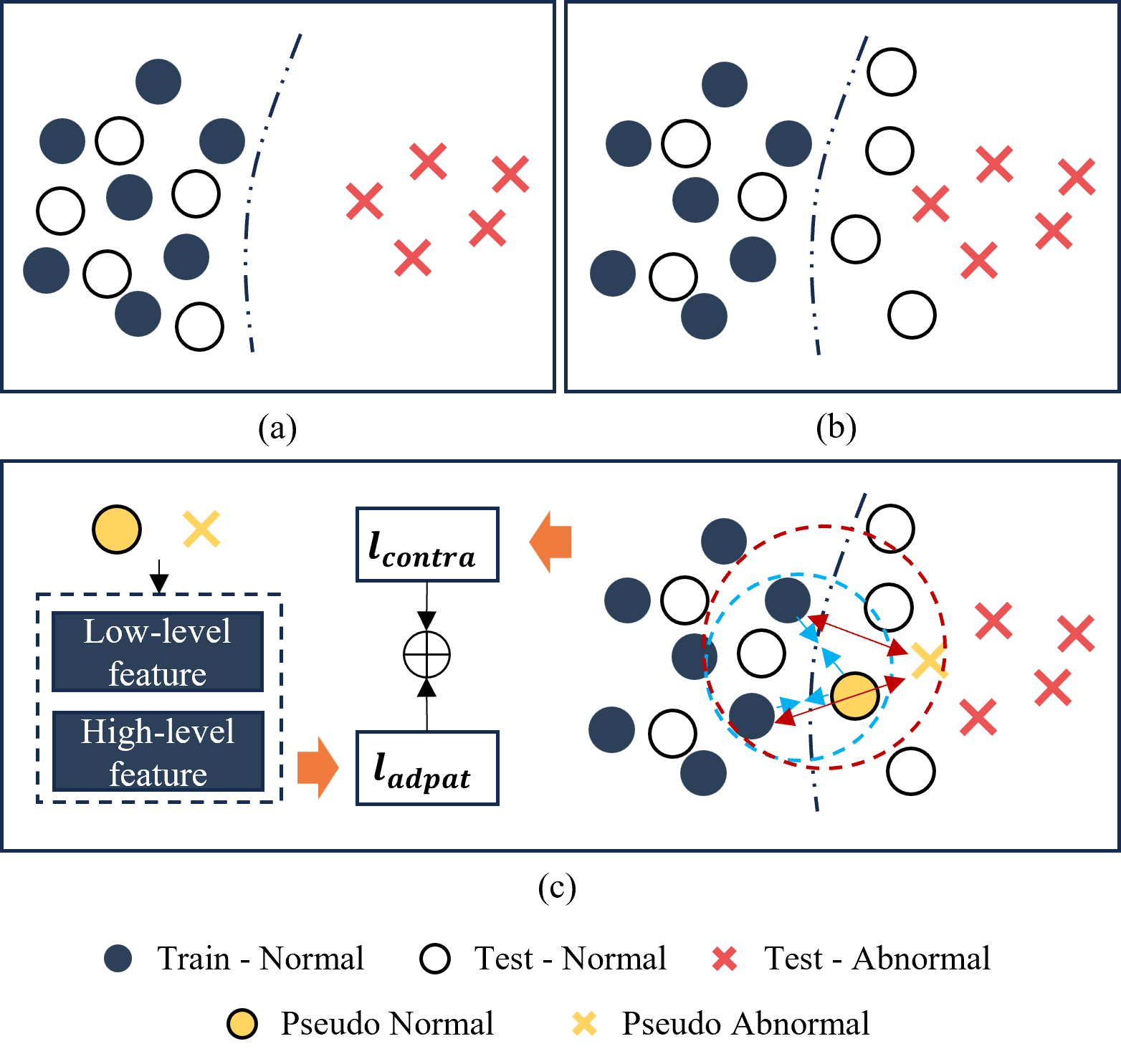}
    \caption{Two cases of the sample distribution. 
    (a) The normality in the test set is consistent with that in the training set.   
    (b) The normality shift occurs between the training and test sets.
    (c) Our framework mitigates normality shifts by learning multi-level normal representations and performing selective, risk-aware adaptation at test time.}
    \label{fig:problem}
    % \vspace{-1.6em}
\end{figure}

However, in practical applications, this static assumption is frequently violated. Due to the limited scale and diversity of available training data, models often capture only a subset of the true normal manifold, resulting in an incomplete characterization of normal patterns. Consequently, when previously unseen but valid normal variations emerge at test time, a phenomenon referred to as normality shift, these representationally deficient models exhibit extreme vulnerability. Because their learned boundaries lack resilience, they erroneously flag these shifted normal instances as anomalies, leading to severe performance degradation and an unacceptable surge in false alarms. As illustrated in~\cref{fig:problem}, while existing unsupervised methods perform reliably under static distributions (Case a), they fail drastically when confronting normality shifts (Case b).
To mitigate the impact of normality shifts, a natural intuition is to adapt the model directly to the unlabeled test data. While test-time adaptation (TTA) strategies offer a potential direction, applying them to unsupervised anomaly detection creates a severe dilemma. First, due to the lack of training supervision, the model's initial characterization of normal patterns is often representationally deficient, leaving it highly susceptible to misguidance during adaptation. Second, and more critically, test data inherently contains unlabeled anomalies. If adaptation is performed indiscriminately across all test samples, it inevitably triggers anomaly contamination. The model blindly absorbs these anomalous patterns, rapidly and irreversibly losing its discriminative capability. Thus, achieving safe adaptation under normality shifts remains a critical bottleneck.

To address these challenges, we propose \textbf{RTTAD}, a \underline{R}isk-aware \underline{T}est-time adaptation framework for unsupervised \underline{T}abular \underline{A}nomaly \underline{D}etection. 
Unlike prior works that treat adaptation as an isolated post-hoc patch, RTTAD holistically manages distribution shifts through a synergistic two-stage paradigm. 
Specifically, as illustrated in~\cref{fig:problem} (c), RTTAD tackles the first challenge by introducing a Collaborative Dual-task Training module. 
By synergizing a main feature reconstruction task with an auxiliary latent-embedding task, the model extracts multi-level feature representations. 
This establishes a comprehensive characterization of normality, acting as a robust anchor that reduces the risk of being misled when adapting to shifted test distributions.
To resolve the second challenge, RTTAD employs a Test-Time Contrastive Learning (TTCL) module. 
Rather than performing indiscriminate adaptation across all test samples, TTCL executes a rigorous risk-aware protocol: it selectively updates the model using high-confidence pseudo-normal samples while explicitly suppressing the influence of potentially anomalous ones. 
Furthermore, TTCL incorporates a $k$-nearest neighbor (KNN)-based contrastive objective to refine localized embedding distributions by pulling pseudo-normal samples tightly toward established normal patterns while pushing anomalies away, effectively mitigating anomaly contamination and bolstering the model's intrinsic discriminative capability.
Extensive experiments on 15 tabular datasets demonstrate that RTTAD consistently achieves state-of-the-art overall detection performance.

Our main contributions can be summarized as follows:
\begin{itemize}
    \item We identify and formalize the phenomenon of normality shifts in unsupervised tabular anomaly detection, elucidating the inherent dilemma between mitigating distribution shifts and the critical risk of anomaly contamination during test-time adaptation.
    \item We propose RTTAD, a synergistic framework that bridges the training and testing phases. It establishes robust normal priors via collaborative dual-task training and performs safe, risk-aware model updates through test-time contrastive learning.
    \item Comprehensive evaluations on 15 tabular datasets with synthetic normality shifts demonstrate the exceptional robustness and effectiveness of the proposed method.
\end{itemize}

\section{Related Work}
\subsection{Unsupervised Tabular Anomaly Detection}
Unsupervised anomaly detection, which does not rely on anomaly labels during the training phase, is one of the most practical approaches to tabular anomaly detection.
Existing studies typically learn the feature patterns of normal samples during training and subsequently identify samples that deviate from the learned patterns as anomalies during testing.
These methods can be broadly classified into four categories: 
One-class classification-based methods~\cite{scholkopf1999support,tax2004support,ruff2018deep,goyal2020drocc,massoli2021mocca,xu2024calibrated} learn a decision boundary that encloses the normal samples, classifying those that fall outside this boundary as anomalies during testing.
Clustering/feature-distribution-based methods~\cite{breunig2000lof,zong2018deep,liu2022unsupervised,ali2024anomaly,li2022ecod} detect anomalies by estimating the density of data points or evaluating their positions within the feature distribution.
Reconstruction-based methods~\cite{schlegl2017unsupervised,schlegl2019f,gong2019memorizing,zavrtanik2021draem,zaheer2022generative,zhang2023unsupervised,guo2024recontrast} learn compact embeddings to model normal feature patterns and classify samples with high reconstruction errors as anomalies.
Self-supervised learning-based methods~\cite{bergman2020classification,qiu2021neural,shenkar2022anomaly,yin2024mcm} design auxiliary tasks to uncover latent data structures and patterns; samples that fail these tasks at test time are flagged as anomalies.

Although existing methods have demonstrated strong performance, they generally assume consistent normality between the training and test sets. 
When this assumption is violated by normality shifts, their detection performance can degrade substantially.
In contrast, our method explicitly accounts for potential normality shifts.
Rather than treating all deviations from the known distribution as anomalies, we introduce a risk-aware test-time adaptation framework that selectively adapts the model. This fundamental difference allows our approach to maintain robust detection performance under distributional changes.
\begin{figure*}[h]
    \centering
    \includegraphics[width=\textwidth]{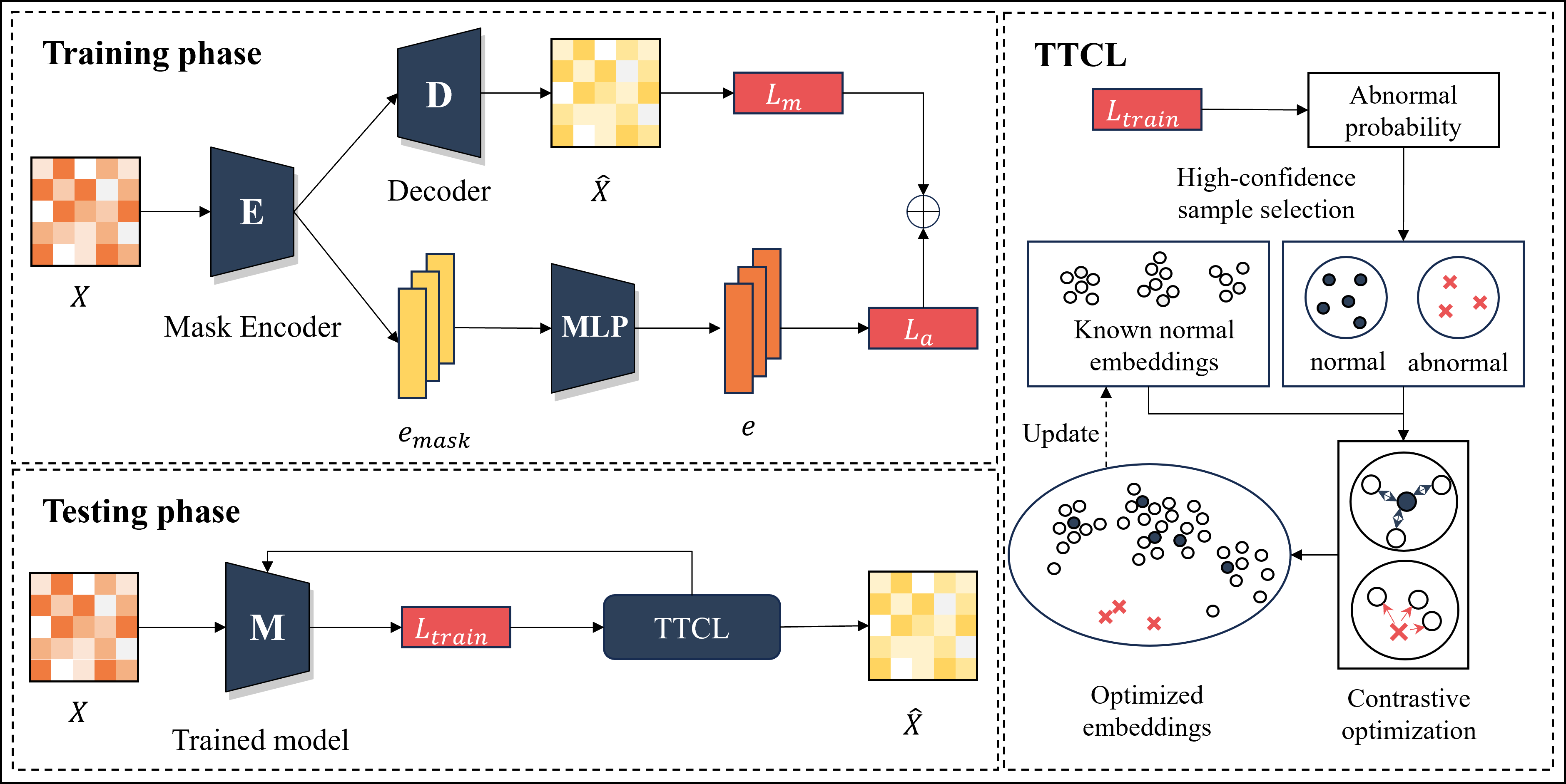}
    \caption{The framework of RTTAD.
    During the training phase, RTTAD employs collaborative dual-task learning to capture multi-level representations of normal samples. Specifically, the primary task reconstructs input features to extract low-level information, while the auxiliary task reconstructs latent embeddings to model high-level feature patterns. During the testing phase, the trained model is iteratively refined via the Test-Time Contrastive Learning (TTCL) module to achieve risk-aware adaptation. Specifically, TTCL selectively utilizes high-confidence pseudo-labels to guide differentiated adaptation strategies for pseudo-normal and pseudo-anomalous samples. Concurrently, a $k$-nearest neighbor (KNN)-based contrastive optimization is employed to further pull the representations of pseudo-normal samples toward known normal patterns while pushing pseudo-anomalous representations away, thereby enhancing the model's discriminative capability between normal and anomalous instances.
    }
    \label{fig:framework}
\end{figure*}

\subsection{Test-time Adaption}
Test-time adaptation~\cite{liu2021subtype,liu2021ttt++,shu2022test,jang2023test,kim2024model} aims to mitigate performance degradation under distribution shifts by enabling pre-trained models to adapt to unlabeled test data dynamically. 
In the realm of tabular anomaly detection, recent works such as TTAD~\cite{cohen2023boosting} and EPHAD~\cite{patra2025evidence} have emerged as preliminary explorations of TTA. 
Specifically, TTAD generates augmented instances by retrieving the $k$-nearest neighbors of test samples and aggregates their predictions to alleviate shift-induced bias; meanwhile, EPHAD adopts a post-hoc calibration strategy, leveraging auxiliary classical detectors to dynamically adjust the outputs of the initial model during the testing phase, thereby reducing misclassifications.

Although these methods have achieved certain effectiveness in specific scenarios, they still exhibit several limitations. First, TTAD's neighborhood aggregation strategy is susceptible to assimilation by surrounding dominant normal instances when anomalous samples are sparse, leading to a degradation in the model's anomaly detection capability. Second, the post-hoc calibration of EPHAD relies on the robustness of the initial detector. If the initial model suffers from an incomplete characterization of normal patterns, such limited post-hoc adjustments may fail to reverse the collapse of the model's basic discriminative capability. More importantly, by treating adaptation merely as an isolated testing-phase patch, they overlook the intrinsic connection between the training and testing stages, making it difficult to effectively resolve the dilemma of anomaly contamination.
In contrast, RTTAD tightly integrates the construction of a robust prior during the training phase with risk-aware contrastive optimization during the testing phase. It adapts to distribution shifts while explicitly suppressing anomaly contamination, effectively preserving the model's anomaly detection capability.

\section{Methodology} \label{sec:method}
\subsection{Problem Statement} \label{sec:method.problem}
In this paper, we focus on unsupervised tabular anomaly detection under potential distributional changes. Given a training set $\mathcal{D}_{train}=\{x_{i}^{train}\}_{i=1}^{N}$ and a test set $\mathcal{D}_{test}=\{x_{i}^{test}\}_{i=1}^{N^{\prime}}$, where $N$ and $N^{\prime}$ represent the number of training and test samples respectively. The test set $\mathcal{D}_{test}$ is composed of unknown normal samples and a small fraction of anomalous samples, with $\alpha$ denoting the contamination rate. 

Traditionally, an unsupervised anomaly detection model $M$ is trained on $\mathcal{D}_{train}$ to learn the feature patterns of normal samples. The trained model $M_{train}$ is then employed to predict the anomaly probabilities $P_{abnormal}$ of test samples. This static process can be formulated as follows:
\begin{equation} M_{train}=M(\mathcal{D}_{train}), \end{equation}
\begin{equation} P_{abnormal}=M_{train}(\mathcal{D}_{test}), \end{equation}
\begin{equation} \mathcal{A} = \{x{i}^{test} | p_{i} \in Top(P_{abnormal}, \alpha), x_{i}^{test} \in \mathcal{D}_{test}\}, \end{equation}
where $Top(P_{abnormal},\alpha)$ denotes the subset of instances with the highest anomaly probabilities corresponding to the top $\alpha\%$. 
However, existing methods typically assume that the distribution of normal samples in the test set is consistent with that in the training set. In practical scenarios, this assumption is often violated due to shifts in normality. Such shifts cause the static model $M_{train}$ to misclassify shifted normal samples as anomalies. 

To alleviate this issue, we consider a test-time adaptation setting. Before generating the final predictions, the model $M_{train}$ is further refined using the unlabeled test data $\mathcal{D}_{test}$ to obtain an updated model $M_{update}$. The objective is to enable the model to capture the evolved normal patterns while explicitly preventing the contamination of anomalous samples during the adaptation process. 

\subsection{Overview of the Proposed RTTAD} \label{sec:method.overview}
RTTAD holistically coordinates model learning across both training and testing phases. 
By establishing robust normal priors during training and executing selective, risk-controlled updates during testing, the framework achieves effective adaptation to distribution shifts while explicitly suppressing anomaly contamination.
Specifically, as illustrated in~\cref{fig:framework}, RTTAD leverages collaborative dual-task learning during the training phase to enhance the representation of normal samples. By capturing multi-level feature patterns, the model builds a comprehensive characterization of normality, which serves as a robust anchor to reduce the risk of being misled by shifted distributions in subsequent stages.  
During the testing phase, the Test-Time Contrastive Learning module enables adaptive model updates through iterative optimization. In this process, TTCL leverages high-confidence pseudo-labels to guide the adaptation, executing differentiated update strategies for pseudo-normal and pseudo-anomalous samples to mitigate the detrimental impact of latent anomalies on the model's detection capability. 
Furthermore, by conducting $k$-nearest neighbor-based contrastive optimization in the embedding space, RTTAD pulls pseudo-normal samples toward the known normal distribution while pushing pseudo-anomalous samples away, which further enhances the model's ability to discriminate between normal and anomalous instances. By coordinating these two stages, RTTAD effectively tackles normality shifts and and enhances the overall anomaly detection performance of the model.

\subsection{Collaborative Dual-task Training} \label{sec:method.train}
The capability of a model to accurately characterize normal patterns relies on the richness of the informative features extracted during training. 
To establish a rich characterization of normal features under the constraints of limited scale and diversity in training data, RTTAD introduces a collaborative dual-task training module. 
Specifically, the implementation of this module entails two synergistic reconstruction tasks. 
First, it executes a primary feature reconstruction task on masked inputs to capture the low-level feature information of normal samples. Second, it introduces an auxiliary latent-embedding reconstruction task to further model high-level feature representations. 
By comprehensively profiling normal patterns across multiple levels, this design constructs a robust representational anchor, providing a solid foundational capability for the subsequent risk-aware adaptation during the testing phase.

\subsubsection{\textbf{Backbone Model}}
The backbone of the model is a masked autoencoder.
Give the input $\textbf{X}\in \mathbb{R}^{B\times d}$ from the training set $\mathcal{D}_{train}$, $B$ is the batch size, $d$ denotes the dimension of the feature vector.
The input $\textbf{X}$ is first passed through the masked encoder $E$, which serves as a shared feature extractor for both tasks. 
This mask encoder $E$ consists of two components: a mask generator $g_{1}$ and an encoder $g_{2}$, $E=g_{1}+g_{2}$. 
$g_{1}$ produces multiple mask tensors $\textbf{X}_{mask} = g_{1}(\textbf{X})$ of the same size as $\textbf{X}$, and leverage a sigmoid function to scale each value of $\textbf{X}_{mask}$ between 0 and 1.
Element-wise multiplication is then applied between $\textbf{X}_{mask}$ and $\textbf{X}$, the obtained masked input is subsequently passed into $g_{2}$ to obtain the masked representation $\textbf{e}_{mask} = E(\textbf{X}) = g_{2}(\textbf{X}_{mask}\odot \textbf{X})$ in the embedding space.

Furthermore, to capture a broader spectrum of information from normal samples, we ensure sufficient diversity in the masking patterns.
This is essential, as using similar masks may cause the model to learn redundant features, which not only fail to improve anomaly detection performance but may also degrade it.
Inspired by MCM~\cite{yin2024mcm}, the diversity of masking patterns is promoted by incorporating a dedicated loss function, as defined in~\cref{eq:diversity_loss},
\begin{equation}
\mathcal{L}_{\text {div }}=\sum_{i=1}^T\left[\ln \left(\sum_{j=1}^T\left(\mathbb{I}_{i \neq j} \cdot e^{\frac{<\textbf{X}_{mask}^{i}, \textbf{X}_{mask}^{j}>}{\tau}} \right)\right) \cdot s \right],
    \label{eq:diversity_loss}
\end{equation}
where $< >$ denotes the inner product operation, $\mathbb{I}_{i \neq j}$ is the indicator function, if $i=j$, $\mathbb{I}_{i \neq j} = 0$, otherwise $\mathbb{I}_{i \neq j}=1$, $\tau$ is a temperature parameter, and $s$ is a scaling factor to adjust the range of the diversity loss, $T$ denotes the number of masks.

\subsubsection{\textbf{Main task: learning low-level features}}
In the main task, the masked representation is fed into the decoder $D$ to reconstruct the original input $\textbf{X}$, as shown in $\hat{\textbf{X}}=D(\textbf{e}_{mask})$. 
By minimizing the reconstruction loss $\mathcal{L}_{m}$ between the input and its reconstruction, the model then learns low-level feature representations of the tabular data. 
The loss function is formulated as~\cref{eq:main_task_Loss}:
\begin{equation}
    \mathcal{L}_{m} = \frac{1}{T}\sum_{i=1}^{T}\| \hat{\textbf{X}}_{i} - \textbf{X} \|^{2}.
    \label{eq:main_task_Loss}
\end{equation}

\subsubsection{\textbf{Auxiliary task: capturing high-level features}}
In the auxiliary task, the masked representation $\mathbf{e}_{mask}$ is fed into a multi-layer perceptron (MLP) to reconstruct the embedding $\textbf{e}$ of the unmasked input. 
By minimizing the reconstruction loss between the predicted and original embeddings, the model captures the intrinsic knowledge embedded in the encoded representations, thereby learning high-level feature representations of the data.
To ensure that $\textbf{e}$ and $\textbf{e}_{mask}$ have the same size, we replicate $\textbf{X}$ $T$ times to match the size of $\textbf{X}_{mask}$, and then pass the replicated input through the encoder $g_{2}$ to obtain $\textbf{e}=g_{2}(\textbf{X}_{T})$, $\textbf{X}_{T}$ represents the input $\textbf{X}$ that has been replicated $T$ times.
The auxiliary task is trained by minimizing the reconstruction loss $\mathcal{L}_{a}$
between the predicted embedding $\hat{\textbf{e}}=MLP(\textbf{e}_{mask})$ and the embedding $\textbf{e}$.
The loss function is formulated as~\cref{eq:auxiliary_task_Loss}:
\begin{equation}
    \mathcal{L}_{a} = \frac{1}{T}\sum_{i=1}^{T}\| \hat{\textbf{e}}_{i} - \textbf{e} \|^{2}
    \label{eq:auxiliary_task_Loss}
\end{equation}

\subsubsection{\textbf{Model Training Loss}}
The overall training loss of the model integrates the reconstruction losses from the main and auxiliary tasks, as well as the mask diversity loss, and is formally defined as~\cref{eq:training_loss},
\begin{equation}
    \mathcal{L}_{Train}= \mathcal{L}_{m} + \lambda \mathcal{L}_{a} + \gamma \mathcal{L}_{div},
    \label{eq:training_loss}
\end{equation}
where $\lambda$ and $\gamma$ are the weights used to adjust the overall loss function.

\subsection{Test-Time Contrastive Learning} \label{sec:method.update}
In unsupervised test-time adaptation, the absence of ground-truth labels and the inevitable presence of anomalies within the test data make indiscriminate model updates highly susceptible to anomaly contamination, which can severely compromise the model's fundamental discriminative capability. To mitigate this risk, the Test-Time Contrastive Learning (TTCL) module is designed to establish a safe and risk-aware adaptation mechanism. Specifically, the implementation of TTCL entails two key strategies. First, it filters high-confidence pseudo-normal and pseudo-anomalous samples based on the initial predictions to guide the adaptation process, executing differentiated updates to avert interference from ambiguous instances. Second, it introduces a $k$-nearest neighbor (KNN)-based contrastive optimization objective in the embedding space. By explicitly pulling pseudo-normal samples toward the established normal anchors and pushing pseudo-anomalous instances away, it further refines the feature representations. This design ensures that the model can adapt to normality shifts while effectively preserving and solidifying the decision boundary between normal and anomalous patterns.

\subsubsection{\textbf{High-Confidence Samples Selection}}
Given the test set $\mathcal{D}_{test}$, we first apply the trained model to output the losses for all test samples and normalize them into the range $[0, 1]$.
Then the normalized losses of test samples can be regarded as their anomaly probability, as shown in~\cref{eq:abnormal_prob}, $Norm$ denotes the Min-Max Scaler.
\begin{equation}
    P_{abnormal} = Norm(M_{train}(\mathcal{D}_{test}))
    \label{eq:abnormal_prob}
\end{equation}
Subsequently, TTCL selects the most confident normal and abnormal samples from the test set based on sorted anomaly scores, referring to them as pseudo-normal and pseudo-anomalous samples.
Here, $\gamma$GMM~\cite{perini2023estimating} is employed to adaptively estimate the confidence threshold, thereby eliminating the requirement for prior knowledge regarding the dataset's contamination rate.
The selected samples can be denoted as~\cref{eq:h_normal,eq:h_abnormal}, 
\begin{equation}
    \mathcal{H}_{normal} = \{ h_{i}^{normal}\}_{i=1}^{C_{normal}}
    \label{eq:h_normal}
\end{equation}
\begin{equation}
    \mathcal{H}_{abnormal} = \{ h_{i}^{abnormal}\}_{i=1}^{C_{abnormal}}
    \label{eq:h_abnormal}
\end{equation}
where $\mathcal{H}_{normal}$ represents the set of high-confidence normal samples and $\mathcal{H}_{abnormal}$ represents the set of high-confidence abnormal samples, $C_{normal}$ and $C_{abnormal}$ denote the number of samples in two sets.

\subsubsection{\textbf{Risk-Aware Model Adaptation}}
At test time, TTCL adapts the model using both the main and auxiliary tasks on the selected samples, without requiring any test labels and thus avoiding label leakage. 
The adaptation process treats pseudo-normal and pseudo-abnormal samples differently: for pseudo-normal samples, the model is encouraged to learn their representations and reduce reconstruction errors to prevent false positives; for pseudo-abnormal samples, the adaptation is constrained to hinder accurate representation learning, ensuring that they retain high reconstruction errors and remain distinguishable as anomalies. 
The overall adaptation objective is summarized in~\cref{eq:adapt_loss}:
\begin{equation}
\begin{aligned}
    \mathcal{L}_{adapt} &= \sigma_s \cdot \frac{1}{C_s} \sum_{i=1}^{C_s} \left( 
\mathcal{L}_m(h_i^{s}) + \lambda \mathcal{L}_a(h_i^{s}) + \gamma \mathcal{L}_{\text{div}} 
\right),\\ \sigma_s &= 
\begin{cases}
+1, & s = \text{normal} \\
-1, & s = \text{abnormal}
\end{cases}
\end{aligned}
\label{eq:adapt_loss}
\end{equation}

\subsubsection{\textbf{Embedding Contrastive Optimization}}
TTCL further refines the representations of samples in embedding space by encouraging pseudo-normal samples to move closer to known normal patterns and pushing pseudo-anomalous samples away.
Rather than contrasting against all known normal samples, which is inefficient and unrealistic due to the multi-pattern nature of normal data, TTCL employs a KNN-based contrastive objective that operates on local neighborhoods. 
This localized formulation improves both discriminative representation learning and computational efficiency. 
The contrastive loss is given in~\cref{eq:contra_loss},
\begin{equation}
\begin{aligned}
    \mathcal{L}_{contra} &= \sigma_s \cdot \frac{1}{C_{s}} \sum_{i=1}^{C_{s}} \| h_{i}^{s} - KNN(h_{i}^{s},\mathcal{O},k) \|^{2} , \\
    \sigma_s &= 
\begin{cases}
+1, & s = \text{normal} \\
-1, & s = \text{abnormal}
\end{cases}
\end{aligned}
\label{eq:contra_loss}
\end{equation}
where $\mathcal{O}$ denotes the embeddings of known normal samples, $KNN(x,\mathcal{O},k)$ denotes finding the $k$-nearest embeddings to the representation of sample $x$ from the set of known normal embeddings $\mathcal{O}$.

\subsubsection{\textbf{Model Update Loss}}
For pseudo-normal or pseudo-anomalous samples, the model jointly optimizes the adaptation loss and the contrastive loss during the update process. 
The overall loss function is defined as~\cref{eq:l_update},
\begin{equation}
    \mathcal{L}_{Update} = \mathcal{L}_{adapt} + \delta \cdot \mathcal{L}_{contra},
    \label{eq:l_update}
\end{equation}
where $\delta$ is a hyperparameter to balance two losses.

\subsubsection{\textbf{Update Iterations}}
Let the initially trained model $M_{train}$ be denoted as $M_{update}^{(0)}$,  and let $n$ denote the total number of update rounds.
At test time, the model is iteratively refined via TTCL. 
Specifically, at round $t$, the model is updated using the current pool of selected normal samples, the updated model can be formulated as~\cref{eq:m_update},
\begin{equation}
    M_{update}^{(t)} = TTCL(M_{update}^{(t-1)};\mathcal{L}_{Update}).
    \label{eq:m_update}
\end{equation}
After each update, newly identified high-confidence normal samples are added to the pool, as shown in~\cref{eq:o_t},
\begin{equation}
    \mathcal{O}^{(t)} = \mathcal{O}^{(t-1)} \cup \mathcal{H}_{normal}^{(t-1)}.
    \label{eq:o_t}
\end{equation}
This iterative process continues until no sufficient samples remain for further selection.
After the final update, the refined model $M_{update}^{(n)}$ is used to compute anomaly scores $P_{test} = M_{update}^{(n)}(\mathcal{D}_{test})$. The predicted label for each test sample is then obtained as shown in~\cref{eq:y_test},
\begin{equation}
    y_{i}^{test} = \mathbb{I} (p_{i}^{test} \ge Percentile(p^{test}, \alpha),
    \label{eq:y_test}
\end{equation}
where $P_{test}$ denotes the anomaly scores of test samples, $y_{i}^{test}$ is the predicted label of sample $i$, $\mathbb{I}(\cdot)$ is the indicator function, and $Percentile(p^{test}, \alpha)$ denotes the $\alpha$-percentile of $p^{test}$.

\section{Experiments} \label{sec:experiment}
\subsection{Normality Shift Construction} \label{sec:data_shift_construction}
Following prior works~\cite{li2022ecod,shenkar2022anomaly,yin2024mcm}, we systematically evaluate our method on 15 commonly used tabular datasets sourced from ODDS~\cite{rayana2016odds} and ADBench~\cite{han2022adbench}.
These datasets span a wide range of domains, scales, feature dimensions, and anomaly ratios, which enhance the generality of our evaluation and strengthens the reliability of the conclusions. 
Detailed statistics of datasets are provided in~\cref{tab:appendix:datasets}.
\begin{table}[h]
    \centering
    \caption{The statistics of datasets.}
    \begin{tabular}{ccccc}
    \hline
    \textbf{Dataset} & \textbf{Samples} & \textbf{Dim} & \textbf{Anomaly} & \textbf{Category}\\ 
    \hline
       Arrhythmia & 452 & 274 & 66 (15\%) & Healthcare \\
       BreastW & 683 & 9 & 239 (35\%) & Healthcare \\
       Cardio & 1831 & 21 & 176 (9.6\%) & Healthcare \\
       Cardiotocography & 2114 & 21 & 466 (22.04\%) & Healthcare \\
       Glass & 214 & 9 & 9 (4.2\%) & Forensic \\
       Ionosphere & 351 & 33 & 126 (36\%) & Oryctognosy \\
       Mammography & 11183 & 6 & 260 (2.32\%) & Healthcare \\
       Optdigits & 5216 & 64 & 150 (2.88\%) & Image \\
       Pendigits & 6870 & 16 & 156 (2.27\%) & Image\\
       Pima & 768 & 8 & 268 (35\%) & Healthcare \\
       Satellite & 6435 & 36 & 2036 (32\%) & Astronautics \\
       Satimage-2 & 5803 & 36 & 71 (1.2\%) & Astronautics \\
       Thyroid & 3772 & 6 & 93 (2.5\%) & Healthcare \\
       Wbc & 278 & 30 & 21 (5.6\%) & Healthcare \\
       % \cdashline{1-5}
       Wine & 129 & 13 & 10 (7.75\%) & Chemistry \\
    \hline
    \end{tabular}
    \label{tab:appendix:datasets}
\end{table}

\begin{figure*}[]
\begin{center}
\includegraphics[width=\textwidth]{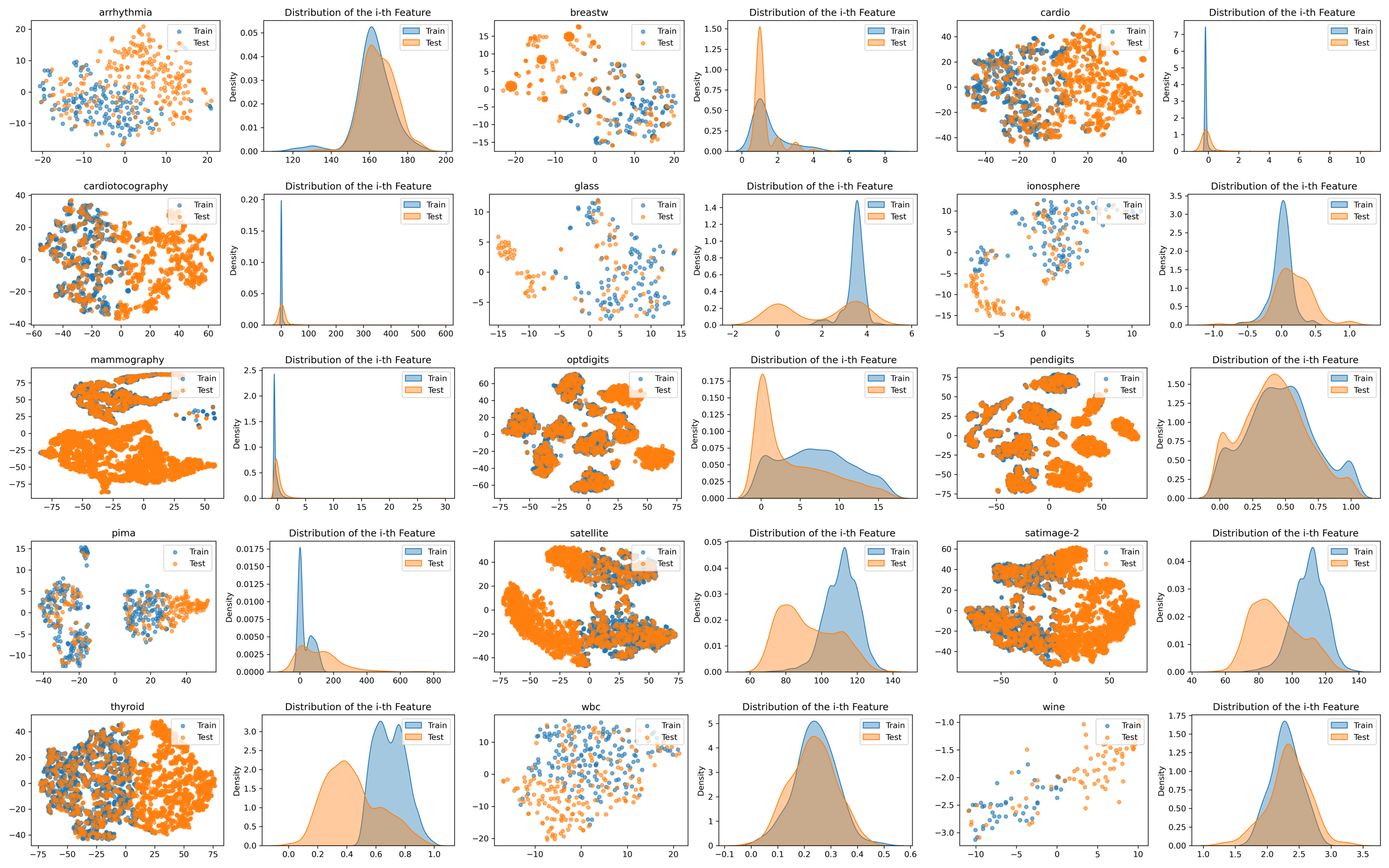}
\end{center}
\caption{Comparison of normal sample distributions between the training and test sets for each dataset, along with the distribution of the $ i$-th feature.}
\label{fig:shift_analysis}
\end{figure*}

Second, to simulate distribution shifts within the normal data, we synthetically construct these shifts by applying K-Means clustering exclusively to the normal samples of each dataset. 
Specifically, samples from the largest cluster are partitioned into both training and test sets, whereas normal samples from the remaining clusters, along with all anomalous samples, are strictly assigned to the test set. 
Consequently, the test set comprises normal samples that conform to the training distribution as well as those that deviate from it. To rigorously validate the rationality of this construction, we adopt the evaluation protocol established in the prior study~\cite{dragoi2022anoshift}, conducting comprehensive qualitative and quantitative analyses. 
This involves utilizing t-SNE visualization, Jeffreys Divergence (JD), and Optimal Transport Dataset Distance (OTDD). Notably, JD quantifies feature-wise distribution discrepancies via normalized histograms, while OTDD captures the geometric divergence between datasets within the original feature space. 
As illustrated in~\cref{fig:shift_analysis}, the normal samples in the test set exhibit a conspicuous distributional shift relative to the training set, a qualitative observation that is further corroborated by the JD and OTDD scores in~\cref{tab:jd&otdd}.

\begin{table}[]
    \caption{The results of Jeffreys Divergence (JD) and Optimal Transport Dataset Distance (OTDD) between the distributions of normal samples in the training and test sets across all datasets.}
    \label{tab:jd&otdd}
    \centering
    \begin{adjustbox}{width=\columnwidth}
    \begin{tabular}{llllllllllllllll}
    \hline
    \textbf{Metrics} & \textbf{arrhythmia} & \textbf{breastw} & \textbf{cardio} & \textbf{cardiotocography} & \textbf{glass} \\ 
    \hline
    JD &  1.39 & 0.48 & 1.72 & 0.46 & 1.28 \\
    OTDD & 0.31 & 0.24 & 0.29 & 0.14 & 0.15 \\ 
    \hline
    \textbf{Metrics} & \textbf{ionosphere} & \textbf{mammography} & \textbf{optdigits} & \textbf{pendigits} & \textbf{pima} \\
    \hline 
    JD & 9.08 & 2.40 & 0.19 & 0.80 & 1.65 \\
    OTDD & 0.39 & 0.09 & 0.55 & 0.49 & 0.16 \\ 
    \hline
    \textbf{Metrics} & \textbf{satellite} & \textbf{satimage-2} & \textbf{thyroid} & \textbf{wbc} & \textbf{wine} \\
    \hline 
    JD & 3.23 & 2.83 & 0.37 & 4.17 & 15.43 \\
    OTDD & 0.01 & 0.33 & 0.24 & 0.25 & 0.27 \\
    \hline
    \end{tabular}
    \end{adjustbox}
\end{table}

\subsection{Experimental Setup} \label{sec:experimental_setup}
\subsubsection{Baselines}
To comprehensively evaluate the proposed approach, we benchmark it against fifteen representative baselines. These comprise thirteen standard anomaly detection methods (IForest, LOF, OCSVM, DeepSVDD, ECOD, GOAD, NeuTral AD, ICL, DIF, SLAD, LUNAR, MCM, DRL) and two recent test-time adaptation approaches (TTAD and EPHAD).
The core concepts of these methods are briefly summarized as follows:
\begin{itemize}
    \item \textbf{IForest}~\cite{liu2008isolation} isolates anomalies by recursively partitioning the data using random splits. 
    The core idea is that anomalies are easier to isolate due to their distinctiveness, requiring fewer partitions compared to normal data points, and this isolation process is used to identify anomalies.

    \item \textbf{LOF}~\cite{breunig2000lof} evaluates the local density of data points by comparing the density of a point with that of its neighbors. 
    Points with significantly lower density than their neighbors are considered anomalies, as they deviate from the expected local structure of the data.

    \item \textbf{OCSVM}~\cite{scholkopf1999support} constructs a hyperplane in a high-dimensional space that maximizes the margin around the normal data. 
    This results in the majority of data points being mapped within the boundary, while points that deviate significantly from this boundary are identified as anomalies.

    \item \textbf{DeepSVDD}~\cite{ruff2018deep} learns a deep feature representation of the data while simultaneously minimizing the volume of a hypersphere that encloses the normal data. 
    Data points that lie outside this learned hypersphere are detected as anomalies.

    \item \textbf{ECOD}~\cite{li2022ecod} leverages the empirical cumulative distribution function (ECDF) to detect anomalies. 
    For each feature in the dataset, ECOD computes the ECDF, which captures the data's distributional properties in a robust and interpretable manner. 
    Points that fall in the extreme tails of the distribution are assigned higher anomaly scores.
        
    \item \textbf{GOAD}~\cite{bergman2020classification} generalizes the class of transformation functions to include affine transformation which allows it to generalize to non-image data. 
    By applying these transformations to the input data, GOAD trains a classifier to distinguish between the transformed versions. 
    At test time, normal data will exhibit predictable patterns under these transformations, while abnormal data fails to conform to these patterns, making it easier to be identified.
        
    \item \textbf{NeuTral AD}~\cite{qiu2021neural} learns a set of neural transformations, parameterized by neural networks, which map the input data to various transformed spaces and capture the intrinsic structure of normal data.
    During the testing phase, samples that do not follow the learned patterns are detected as anomalies.

    \item \textbf{ICL}~\cite{shenkar2022anomaly} employs contrastive loss to learn mappings that maximize the mutual information between each sample and the part that is masked out and capture the structure of the samples of the single training class.
    Test samples are scored by measuring whether the learned mappings lead to a small contrastive loss using the masked parts of this sample.
    Samples with high loss values are regarded as anomalies.

    \item \textbf{DIF}~\cite{xu2023deep} uses randomly initialized neural networks to create random representation ensembles. 
    Through random axis-parallel cuts on these representations, it realizes nonlinear partitioning in the original space. 
    With CERE for efficient feature mapping and DEAS combining path length and feature deviation, DIF scores anomalies via isolation tree ensembles.

    \item \textbf{SLAD}~\cite{xu2023fascinating} introduces scale learning for tabular anomaly detection, defining "scale" as the dimensionality relationship between data sub-vectors and their representations. 
    It uses a neural network to learn distribution alignment of subspace transformations via Jensen-Shannon divergence loss, modeling inlier structural regularities. 
    Test instances are scored by divergence from learned scale distributions, high loss indicates anomalies.

    \item \textbf{LUNAR}~\cite{goodge2022lunar} reframes local outlier detection as a GNN message-passing problem, where samples are nodes connected to k-nearest neighbors. It replaces fixed aggregation rules with learnable neural aggregation and trains with synthetic negatives, enabling adaptive, robust anomaly detection.

    \item \textbf{TTAD}~\cite{cohen2023boosting} reduces prediction bias in tabular test-time augmentation by retrieving a test sample's nearest neighbors via an adaptive Siamese distance metric, and utilizing SMOTE or k-Means centroids to generate diverse and in-distribution augmented samples.
    
    \item \textbf{EPHAD}~\cite{patra2025evidence} mitigates the performance degradation caused by data contamination by dynamically calibrating anomaly scores in a post-hoc manner, combining the prior scores of a pre-trained model with auxiliary test-time evidence such as classical anomaly detectors.
    
    \item \textbf{MCM}~\cite{yin2024mcm} adapts mask modeling to address the problem of tabular data anomaly detection.
    Mask generator and autoencoder are employed to capture intrinsic correlations between features existing in training tabular data and model the “characteristic patterns” by such correlations.
    Samples that deviate from these correlations are predicted as anomalies.

    \item \textbf{DRL}~\cite{ye2025drl} tackles tabular anomaly detection by mapping data into a constrained latent space, where each normal sample is represented as a weighted linear combination of fixed orthogonal basis vectors. It enhances discriminability by increasing the variance of normal weights and preserves feature correlations via alignment loss. 
    
\end{itemize}

\subsubsection{Evaluation Metrics}
Following established evaluation protocols in recent literature~\cite{shenkar2022anomaly, yin2024mcm, ye2025drl}, we employ AUC-PR, AUC-ROC, and the F1-score to comprehensively assess the detection performance.

\subsubsection{Implementation details}
Implementation Details. All models are implemented in PyTorch~\cite{paszke2019pytorch} and executed on a single NVIDIA GeForce RTX 2080 Ti GPU. During the training phase, the models are optimized for 200 epochs using the Adam optimizer with a batch size of 512 and a weight decay of $1 \times 10^{-5}$. We employ the ExponentialLR scheduler with a decay factor of 0.98. During inference, the parameter $k$ is set to 3 across all datasets. To adaptively balance the weights between the main and auxiliary tasks across varying datasets, the hyperparameter $\lambda$ is dynamically set to $\min(1.0, 1.0 / \mathcal{L}_{m})$. The hyperparameter $\gamma$ follows the configuration in MCM~\cite{yin2024mcm}, and $\delta$ is set to 1 by default. For the baseline methods, iForest, LOF, OCSVM, DeepSVDD, ECOD, and LUNAR are implemented via the PyOD library~\cite{zhao2019pyod}, while DIF, GOAD, NeuTral AD, ICL, and SLAD utilize the DeepOD library~\cite{xu2023deep,xu2024calibrated}. TTAD, EPHAD, MCM and DRL are executed using their official open-source implementations. To ensure statistical reliability, the reported results for the main and ablation experiments are averaged over three independent runs, whereas the remaining experiments report single-run outcomes.

\begin{table*}[htbp]
\centering
\caption{Comparison of AUC-PR($\uparrow$)results between baseline methods and RTTAD on 15 datasets.}
\resizebox{\textwidth}{!}{
\begin{tabular}{l|*{15}{c}|c}
\hline
 & Iforest & LOF & OCSVM & DeepSVDD & ECOD & GOAD & NeuTraLAD & ICL & DIF & SLAD & LUNAR & TTAD & EPHAD & MCM & DRL & RTTAD \\
\hline
arrhythmia & 0.6019 & 0.5676 & 0.6111 & 0.6115 & 0.6244 & 0.5867 & 0.5023 & 0.5407 & 0.6294 & 0.5372 & 0.5856 & 0.5519 & 0.6411 & 0.5657 & 0.5510 & 0.6212 \\
breastw & 0.8536 & 0.7818 & 0.7656 & 0.8256 & 0.7581 & 0.9860 & 0.5662 & 0.8508 & 0.9737 & 0.9569 & 0.9644 & 0.9836 & 0.8588 & 0.9911 & 0.9302 & 0.9921 \\
cardio & 0.5381 & 0.5581 & 0.5835 & 0.4407 & 0.6860 & 0.4606 & 0.2458 & 0.3687 & 0.6176 & 0.4277 & 0.4782 & 0.6156 & 0.5846 & 0.6849 & 0.4054 & 0.6385 \\
cardiotocography & 0.5529 & 0.5562 & 0.6531 & 0.5417 & 0.6120 & 0.3408 & 0.3746 & 0.4113 & 0.4944 & 0.3255 & 0.4604 & 0.5321 & 0.5155 & 0.4051 & 0.3682 & 0.3906 \\
glass & 0.1721 & 0.3482 & 0.3472 & 0.5118 & 0.1658 & 0.0994 & 0.1201 & 0.2309 & 0.1013 & 0.1105 & 0.5750 & 0.2134 & 0.1204 & 0.1099 & 0.1191 & 0.1504 \\
ionosphere & 0.8094 & 0.8032 & 0.8094 & 0.7772 & 0.6842 & 0.6543 & 0.6337 & 0.6063 & 0.8097 & 0.7038 & 0.8241 & 0.9708 & 0.8170 & 0.7504 & 0.8282 & 0.7552 \\
mammography & 0.2713 & 0.4927 & 0.4582 & 0.4812 & 0.4862 & 0.3293 & 0.0563 & 0.1560 & 0.4507 & 0.1467 & 0.4962 & 0.0861 & 0.3530 & 0.4781 & 0.1022 & 0.5407 \\
optdigits & 0.3311 & 0.5605 & 0.5290 & 0.0348 & 0.0373 & 0.0640 & 0.0528 & 0.0725 & 0.0647 & 0.0690 & 0.5526 & 0.1296 & 0.0538 & 0.1103 & 0.0997 & 0.4199 \\
pendigits & 0.3430 & 0.5283 & 0.5162 & 0.1523 & 0.4147 & 0.0445 & 0.0362 & 0.0431 & 0.5583 & 0.0475 & 0.5288 & 0.0944 & 0.2744 & 0.0430 & 0.0333 & 0.0964 \\
pima & 0.7448 & 0.7441 & 0.7763 & 0.6826 & 0.7113 & 0.5587 & 0.5373 & 0.5787 & 0.5943 & 0.5902 & 0.6611 & 0.5733 & 0.6344 & 0.5707 & 0.5347 & 0.5921 \\
satellite & 0.7169 & 0.7177 & 0.7173 & 0.6894 & 0.6437 & 0.7595 & 0.8306 & 0.7927 & 0.7173 & 0.8339 & 0.6926 & 0.8472 & 0.7020 & 0.8199 & 0.8400 & 0.8307 \\
satimage-2 & 0.5006 & 0.5013 & 0.5094 & 0.5129 & 0.5931 & 0.6819 & 0.8071 & 0.9461 & 0.9754 & 0.9588 & 0.5114 & 0.9552 & 0.5800 & 0.9717 & 0.0925 & 0.9716 \\
thyroid & 0.6202 & 0.4433 & 0.4208 & 0.3091 & 0.5818 & 0.1503 & 0.1919 & 0.1677 & 0.2379 & 0.4751 & 0.5032 & 0.1846 & 0.2834 & 0.2925 & 0.0629 & 0.7773 \\
wbc & 0.6235 & 0.5745 & 0.6175 & 0.5703 & 0.5609 & 0.5047 & 0.2311 & 0.5735 & 0.4594 & 0.2623 & 0.6094 & 0.5239 & 0.4526 & 0.7268 & 0.4460 & 0.8147 \\
wine & 0.5627 & 0.5781 & 0.5610 & 0.5641 & 0.3177 & 0.9909 & 0.9667 & 0.8813 & 0.8112 & 0.9573 & 0.6667 & 0.8967 & 0.1998 & 0.9909 & 0.9430 & 0.9909 \\
\hline
Average PR & 0.5494 & 0.5837 & 0.5917 & 0.5136 & 0.5251 & 0.4808 & 0.4102 & 0.4814 & 0.5664 & 0.4935 & 0.6073 & 0.5439 & 0.4714 & 0.5674 & 0.4238 & \textbf{0.6388} \\
Average Ranking & 7.73 & 7.07 & 7.13 & 9.20 & 8.47 & 10.33 & 12.33 & 10.53 & 7.33 & 10.07 & 6.33 & 7.27 & 8.47 & 7.47 & 11.27 & \textbf{4.87} \\
\hline
\end{tabular}
}
\label{tab:pr_results}
\end{table*}
\begin{table*}[htbp]
\centering
\caption{Comparison of AUC-ROC($\uparrow$)results between baseline methods and RTTAD on 15 datasets.}
\resizebox{\textwidth}{!}{
\begin{tabular}{l|*{15}{c}|c}
\hline
 & Iforest & LOF & OCSVM & DeepSVDD & ECOD & GOAD & NeuTraLAD & ICL & DIF & SLAD & LUNAR & TTAD & EPHAD & MCM & DRL & RTTAD \\
\hline
arrhythmia & 0.7229 & 0.6797 & 0.5 & 0.5022 & 0.7175 & 0.7694 & 0.7127 & 0.7115 & 0.8167 & 0.7227 & 0.7089 & 0.7866 & 0.8248 & 0.7629 & 0.7345 & 0.8190 \\
breastw & 0.8111 & 0.6469 & 0.5973 & 0.7557 & 0.5725 & 0.9814 & 0.6669 & 0.8860 & 0.9640 & 0.9467 & 0.9609 & 0.9779 & 0.7801 & 0.9936 & 0.9660 & 0.9933 \\
cardio & 0.5943 & 0.6497 & 0.6929 & 0.6019 & 0.8388 & 0.6489 & 0.5901 & 0.6479 & 0.9244 & 0.7109 & 0.5633 & 0.8691 & 0.9002 & 0.6849 & 0.6418 & 0.8199 \\
cardiotocography & 0.4837 & 0.4760 & 0.5018 & 0.5480 & 0.6695 & 0.4619 & 0.4829 & 0.4679 & 0.7682 & 0.4554 & 0.4094 & 0.5737 & 0.6032 & 0.6476 & 0.4843 & 0.6402 \\
glass & 0.4249 & 0.5437 & 0.5384 & 0.6466 & 0.5236 & 0.5390 & 0.6194 & 0.6147 & 0.4031 & 0.6017 & 0.7287 & 0.8074 & 0.5525 & 0.6190 & 0.6076 & 0.7021 \\
ionosphere & 0.6587 & 0.6407 & 0.6587 & 0.5501 & 0.5664 & 0.6288 & 0.6342 & 0.6371 & 0.7559 & 0.6885 & 0.7048 & 0.9546 & 0.7177 & 0.7885 & 0.8112 & 0.7446 \\
mammography & 0.4956 & 0.5918 & 0.8010 & 0.5835 & 0.7321 & 0.8618 & 0.6944 & 0.8014 & 0.8561 & 0.7547 & 0.5898 & 0.6479 & 0.8371 & 0.8635 & 0.7139 & 0.8970 \\
optdigits & 0.6219 & 0.7765 & 0.5 & 0.465 & 0.4839 & 0.5949 & 0.4924 & 0.645 & 0.5466 & 0.6240 & 0.7386 & 0.7794 & 0.5088 & 0.8295 & 0.7441 & 0.9460 \\
pendigits & 0.6461 & 0.6945 & 0.8033 & 0.5252 & 0.6298 & 0.6016 & 0.5021 & 0.5763 & 0.9446 & 0.5895 & 0.6666 & 0.7907 & 0.8916 & 0.6447 & 0.4586 & 0.7313 \\
pima & 0.5626 & 0.5662 & 0.5 & 0.4893 & 0.4893 & 0.5769 & 0.5122 & 0.4813 & 0.5084 & 0.5574 & 0.5579 & 0.5324 & 0.5809 & 0.5251 & 0.6127 & 0.4798 \\
satellite & 0.5634 & 0.5694 & 0.5 & 0.6331 & 0.6086 & 0.7155 & 0.7912 & 0.7756 & 0.6738 & 0.8006 & 0.5747 & 0.7911 & 0.6441 & 0.8065 & 0.7992 & 0.8143 \\
satimage-2 & 0.6724 & 0.6866 & 0.5 & 0.8799 & 0.9124 & 0.9640 & 0.9960 & 0.9958 & 0.9973 & 0.9972 & 0.5893 & 0.9936 & 0.9692 & 0.9986 & 0.7863 & 0.9985 \\
thyroid & 0.9196 & 0.5951 & 0.6239 & 0.5618 & 0.8092 & 0.6534 & 0.7277 & 0.6166 & 0.8838 & 0.8444 & 0.6500 & 0.7257 & 0.9298 & 0.7696 & 0.5262 & 0.9630 \\
wbc & 0.8289 & 0.7866 & 0.8372 & 0.7382 & 0.7694 & 0.9032 & 0.7881 & 0.9234 & 0.8877 & 0.8139 & 0.7995 & 0.8940 & 0.8329 & 0.9643 & 0.8887 & 0.9723 \\
wine & 0.7736 & 0.6250 & 0.5 & 0.5278 & 0.5875 & 0.9986 & 0.9931 & 0.9889 & 0.9542 & 0.9944 & 0.8611 & 0.9828 & 0.6241 & 0.9988 & 0.9931 & 0.9986 \\
\hline
Average ROC & 0.6520 & 0.6352 & 0.6036 & 0.6006 & 0.6665 & 0.7223 & 0.6782 & 0.7198 & 0.7956 & 0.7402 & 0.6714 & 0.8071 & 0.7465 & 0.7990 & 0.7090 & \textbf{0.8408} \\
Average Ranking & 10.40 & 10.87 & 11.27 & 12.60 & 10.07 & 8.20 & 10.47 & 9.07 & 6.00 & 7.67 & 10.20 & 5.60 & 8.07 & 3.80 & 8.80 & \textbf{2.73} \\
\hline
\end{tabular}
}
\label{tab:roc_results}
\end{table*}
\begin{table*}[htbp]
\centering
\caption{Comparison of F1($\uparrow$)results between baseline methods and RTTAD on 15 datasets.}
\resizebox{\textwidth}{!}{
\begin{tabular}{l|*{15}{c}|c}
\hline
 & Iforest & LOF & OCSVM & DeepSVDD & ECOD & GOAD & NeuTraLAD & ICL & DIF & SLAD & LUNAR & TTAD & EPHAD & MCM & DRL & RTTAD \\
\hline
arrhythmia & 0.5466 & 0.4842 & 0.3636 & 0.3646 & 0.5691 & 0.5909 & 0.5303 & 0.4697 & 0.5909 & 0.5152 & 0.5325 & 0.5909 & 0.6212 & 0.5 & 0.5 & 0.5455 \\
breastw & 0.8284 & 0.721 & 0.6938 & 0.7888 & 0.6809 & 0.9665 & 0.6360 & 0.7876 & 0.9436 & 0.8852 & 0.9590 & 0.9456 & 0.7322 & 0.9540 & 0.9205 & 0.9582 \\
cardio & 0.2964 & 0.3319 & 0.3628 & 0.3051 & 0.6509 & 0.5170 & 0.2670 & 0.4545 & 0.5909 & 0.4830 & 0.2773 & 0.6023 & 0.5454 & 0.3239 & 0.3864 & 0.6023 \\
cardiotocography & 0.4218 & 0.4203 & 0.4704 & 0.4433 & 0.5435 & 0.3004 & 0.3305 & 0.3004 & 0.5365 & 0.2961 & 0.3374 & 0.4270 & 0.4785 & 0.3584 & 0.3348 & 0.4099 \\
glass & 0.0870 & 0.1724 & 0.1695 & 0.2192 & 0.1250 & 0. & 0. & 0.2222 & 0.1111 & 0. & 0.2609 & 0.1111 & 0.1111 & 0. & 0. & 0.1111 \\
ionosphere & 0.7654 & 0.7561 & 0.7654 & 0.7143 & 0.5398 & 0.5635 & 0.6190 & 0.5952 & 0.6746 & 0.6270 & 0.7871 & 0.8810 & 0.6746 & 0.6349 & 0.7143 & 0.6508 \\
mammography & 0.0664 & 0.0853 & 0.2836 & 0.0839 & 0.1340 & 0.4154 & 0.0038 & 0.2192 & 0.4692 & 0.1577 & 0.0848 & 0.0615 & 0.3615 & 0.4923 & 0.1 & 0.5407 \\
optdigits & 0.1709 & 0.2158 & 0.1095 & 0.0059 & 0.0081 & 0. & 0. & 0.02 & 0.04 & 0. & 0.1905 & 0.1400 & 0.0200 & 0.0200 & 0.0133 & 0.4733 \\
pendigits & 0.1317 & 0.1181 & 0.1906 & 0.0817 & 0.3574 & 0. & 0. & 0.0256 & 0.5256 & 0.0385 & 0.1089 & 0.0449 & 0.3461 & 0.0064 & 0.0192 & 0.1154 \\
pima & 0.6689 & 0.6667 & 0.7118 & 0.5709 & 0.5738 & 0.5746 & 0.5522 & 0.5448 & 0.5821 & 0.5784 & 0.4847 & 0.5896 & 0.6082 & 0.5149 & 0.5373 & 0.5858 \\
satellite & 0.6261 & 0.6286 & 0.6059 & 0.6144 & 0.5171 & 0.6051 & 0.7194 & 0.6685 & 0.5953 & 0.7083 & 0.6102 & 0.7107 & 0.5997 & 0.7141 & 0.7269 & 0.7210 \\
satimage-2 & 0.0561 & 0.0585 & 0.0369 & 0.1723 & 0.4710 & 0.6620 & 0.8592 & 0.9014 & 0.9577 & 0.9014 & 0.0446 & 0.8873 & 0.5492 & 0.9296 & 0.1268 & 0.9296 \\
thyroid & 0.4674 & 0.0975 & 0.1084 & 0.0942 & 0.5660 & 0.1398 & 0.1828 & 0.1613 & 0.1935 & 0.4516 & 0.1093 & 0.1720 & 0.3225 & 0.2903 & 0.0430 & 0.6667 \\
wbc & 0.3962 & 0.3725 & 0.4301 & 0.3077 & 0.5306 & 0.5238 & 0.2381 & 0.6667 & 0.5238 & 0.1905 & 0.3590 & 0.5238 & 0.4285 & 0.6667 & 0.5238 & 0.7143 \\
wine & 0.5 & 0.2703 & 0.2174 & 0.2273 & 0.2727 & 0.9 & 0.9 & 0.9 & 0.6 & 0.9 & 0.5 & 0.8 & 0.3 & 0.9 & 0.9 & 0.9 \\
\hline
Average F1 & 0.4020 & 0.3599 & 0.3680 & 0.3329 & 0.4360 & 0.4506 & 0.3892 & 0.4625 & 0.5290 & 0.4489 & 0.3764 & 0.4992 & 0.4466 & 0.4870 & 0.3898 & \textbf{0.5953} \\
Average Ranking & 8.33 & 9.20 & 9.27 & 10.80 & 8.27 & 9.00 & 11.40 & 8.60 & 6.00 & 9.40 & 9.33 & 6.27 & 7.53 & 7.93 & 9.60 & \textbf{4.20} \\
\hline
\end{tabular}
}
\label{tab:f1_results}
\end{table*}

\subsection{Main Results}
\textbf{RTTAD achieves the state-of-the-art overall detection performance.}
The detailed quantitative results of all evaluated methods in terms of AUC-PR, AUC-ROC, and the F1-score are tabulated in~\cref{tab:pr_results,tab:roc_results,tab:f1_results}, respectively. 
These results demonstrate that the proposed RTTAD yields the most competitive overall detection performance across all 15 datasets.

\begin{figure*}[htbp]
% \vspace{-0.5em}
\begin{center}
\includegraphics[width=\textwidth]{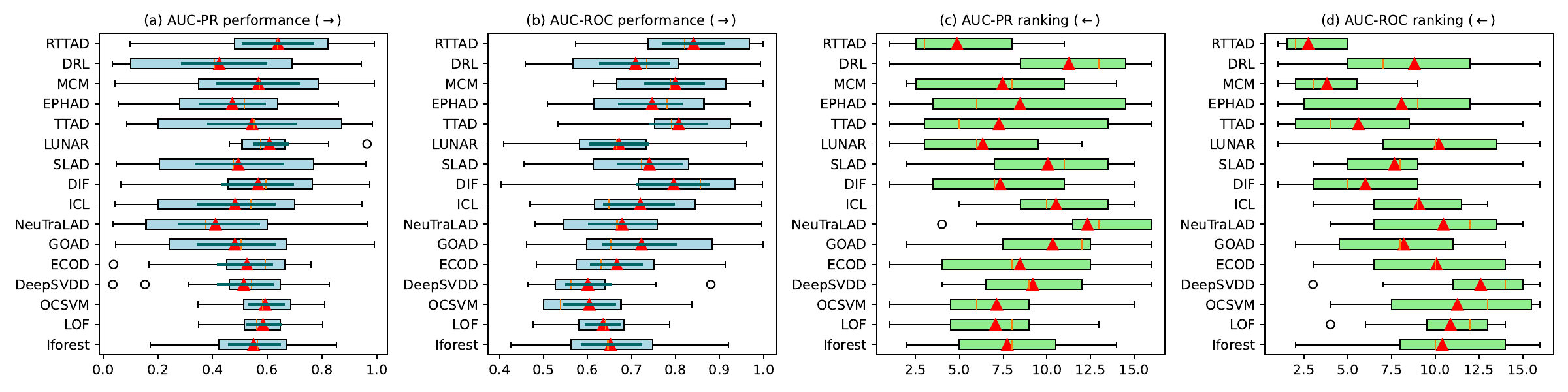}
\end{center}
 % \vspace{-1em}
\caption{Comparison of all models’ performance and ranking across different datasets in terms of AUC-PR
 and AUC-ROC. The triangles represent the average value over all datasets. The dark blue lines in (a) and (b) indicate the confidence intervals of the method’s performance.}
 \label{fig:main_box_pr_roc}
 % \vspace{-1.1em}
\end{figure*}

\begin{figure*}[htbp]
% \vspace{-0.9em}
\begin{center}
\includegraphics[width=\textwidth]{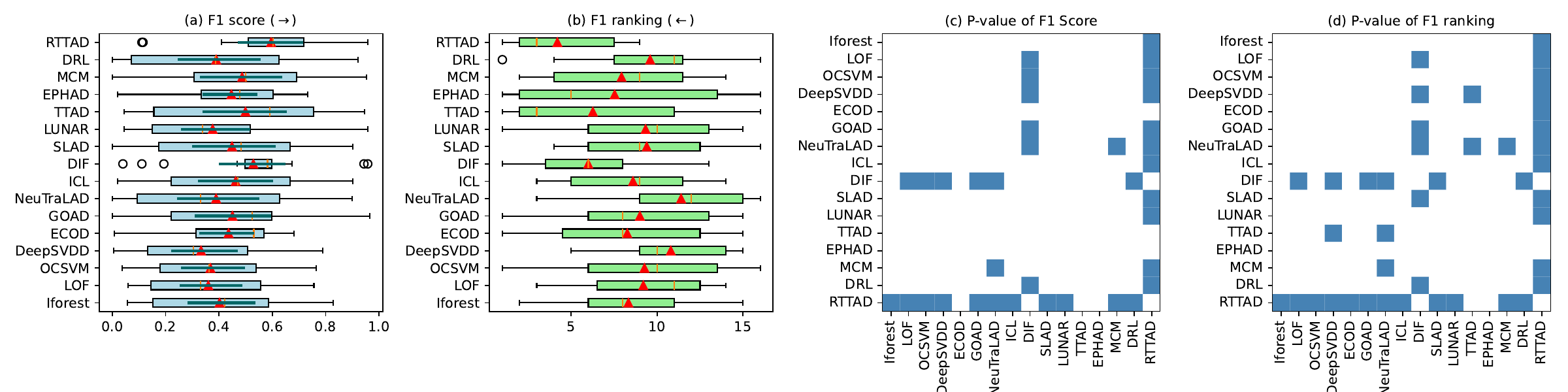}
\end{center}
 % \vspace{-0.2em}
\caption{Comparison of F1 scores. (a) and (b) compare the F1 scores and rankings of all models across different datasets, where the triangles denote the averages over all datasets, and the dark blue lines in (a) indicate the confidence intervals of the method’s performance. (c) and (d) conduct Wilcoxon tests across models and datasets. Blue cells indicate corresponding p-values below 0.05 (significant), while white cells indicate p-values above 0.05 (not significant).}
 \label{fig:main_box_f1}
 % \vspace{-1.1em}
\end{figure*}

To facilitate a straightforward visual comparison of the respective methods, we provide box plots to illustrate the distribution and stability of the detection performance for each approach across all datasets.
\cref{fig:main_box_pr_roc} summarizes the AUC-PR and AUC-ROC results of 16 methods evaluated on 15 datasets. 
Subfigures (a) and (b) present the distributions of AUC-PR and AUC-ROC across datasets, respectively, while subfigures (c) and (d) show the corresponding ranking distributions.
Overall, RTTAD achieves the best average performance across datasets under both metrics. In particular, RTTAD outperforms the strongest baseline by 3.15\% in AUC-PR, with an average ranking improvement of 1.46 positions, and by 3.37\% in AUC-ROC, with an average ranking improvement of 1.07 positions.
\cref{fig:main_box_f1} reports the F1-score performance of different methods. 
Subfigures (a) and (b) show the distribution of F1 scores across all datasets and the corresponding ranking distributions, respectively. 
Overall, RTTAD achieves the best average F1 performance, surpassing the second-best method by 9.61\%, with an average ranking improvement of 2.07 positions.
Subfigures (c) and (d) employ the Wilcoxon signed-rank test~\cite{woolson2007wilcoxon} (with $\alpha$ = 0.05) to assess the statistical significance of the improvements.
Statistical tests reveal that, at a 95\% confidence level, RTTAD achieves statistically significant performance gains compared to 11 out of the 15 baselines.


\textbf{Existing unsupervised tabular anomaly detection methods are inherently vulnerable to normality shifts due to their strict reliance on consistent data distributions.}
Specifically, unsupervised methods rely on the strong assumption of consistent normal patterns between the training and test sets, limiting their detection efficacy in practical scenarios.
Although state-of-the-art detection methods, such as MCM and DRL, have demonstrated exceptional detection capabilities through sophisticated designs in representation learning and feature decoupling, their performance undergoes drastic fluctuations when the normal distribution of the test set deviates from that of the training set. 
In contrast, by synergizing multi-level feature capture during the training phase with a dynamic model adaptation mechanism during the testing phase, RTTAD effectively mitigates the performance degradation induced by normality shifts, thereby achieving superior and more robust overall detection performance.

Test-time adaptation methods exhibit superior overall detection performance compared to unsupervised anomaly detection methods, yet they lack a unified consideration of both the training and testing phases. 
The fact that TTAD and EPHAD generally outperform MCM and DRL across all three evaluation metrics indicates that mitigating normality shifts yields tangible performance gains. 
However, their performance still lags behind that of our proposed RTTAD. 
Specifically, TTAD generates diverse instances by simulating a subset of $k$-nearest neighbors for a given sample and then aggregating their predictions to mitigate shift-induced prediction bias. 
While this strategy proves effective on datasets with high anomaly ratios, it critically falters when anomalies are sparse (e.g., mammography and thyroid). 
In such scenarios, the $k$-nearest neighbor aggregation tends to skew excessively toward normal predictions, masking the detection of latent anomalies and consequently degrading performance. 
EPHAD, which employs post-hoc calibration to adjust the initial detector's predictions, achieves commendable overall performance but remains heavily bottlenecked by the initial detector. When the initial detector suffers severe performance degradation due to distribution shifts on certain datasets, the post-hoc calibration is insufficient to salvage the detection efficacy. 
In contrast, RTTAD leverages multi-level features during the training phase to enhance the model's feature extraction capability, laying a solid foundation for subsequent model adaptation. 
Furthermore, it introduces a risk-aware test-time contrastive learning module during the testing phase, differentially treating latent normal and anomalous samples to avert adaptation failure. 
By holistically coordinating the training and testing phases, RTTAD achieves substantially superior detection performance over both TTAD and EPHAD.

\subsection{Analysis of Pseudo-Label Noise Impact.} 
\begin{table}[htbp]
% \vspace{-1em}
\caption{True rate of pseudo labels in early iterations (true rate of normal labels-true rate of abnormal labels).}
% \vspace{-0.6em}
\label{tab:label_noise}
\begin{center}
% \scalebox{0.7}{
\begin{tabular}{llllll}
    \hline
    \textbf{Dataset} & \textbf{iter 1} & \textbf{iter 2} & \textbf{iter 3} & \textbf{iter 4} \\
    \hline
    pendigits & 1.00-0.07 & 1.00-0.00 & 1.00-0.00 & 1.00-0.00 \\
    cardiotocography & 0.98-0.59 & 0.75-0.52 & 0.96-0.24 & 0.74-0.67 \\
    cardio & 1.00-0.88 & 1.00-0.82 & 1.00-0.76 & 0.91-0.88 \\
    breastw & 1.00-1.00 & 1.00-1.00 & 1.00-1.00 & 1.00-1.00 \\
    \hline
\end{tabular}
% }
\end{center}
\vspace{-1.5em}
\end{table}
\begin{table}[htbp]
% \vspace{-1em}
\caption{Average detection performance across 15 datasets under different forget rates for filtering pseudo-label noise.}
\label{tab:co_teaching}
\begin{center}
% \scalebox{0.7}{
\begin{tabular}{lccc}
    \hline
    \textbf{Forget rate} & \textbf{auc-roc} & \textbf{auc-pr} & \textbf{pr} \\
    \hline
    0\% & 0.8408 & 0.6388 & 0.5953 \\
    10\% & 0.7710 & 0.5829 & 0.5351 \\
    20\% &  0.8201 & 0.6296 & 0.5729 \\
    30\% &  0.7774 & 0.5875 & 0.5318 \\
    40\% &  0.8290 & 0.6630 & 0.6221 \\
    \hline
\end{tabular}
% }
\end{center}
\vspace{-1em}
\end{table}
In the TTCL module of RTTAD, pseudo-labels are assigned to samples with high-confidence predictions. 
To examine whether noisy pseudo-labels lead to persistent performance degradation during test-time adaptation, we select four datasets (pendigits, cardiotocography, cardio, and breastw) with markedly different overall performance and track the accuracy of pseudo-labels during the early adaptation iterations.
The statistical results are summarized in~\cref{tab:label_noise}. 
We observe that pseudo-label accuracy exhibits fluctuations rather than a monotonic decline, indicating that labeling errors are not continuously amplified during adaptation. 
Moreover, even on the pendigits dataset, where RTTAD achieves the weakest performance and the true anomaly rate eventually drops to zero, the final detection results of RTTAD still surpass those of most baseline methods.
These observations suggest that: (1) pseudo-label noise does not cause persistent degradation during test-time adaptation; and (2) despite the presence of noisy pseudo-labels, the benefits of selectively leveraging them outweigh their potential drawbacks in handling normality shifts.

Furthermore, we explore a co-teaching mechanism in which two lightweight MLPs collaboratively select low-loss samples to mitigate pseudo-label noise. As detailed in~\cref{tab:co_teaching}, we evaluate the model under varying forget rates of 10\%, 20\%, 30\%, and 40\%. 
Interestingly, increasing the forget rate does not monotonically improve performance; this is likely attributable to the fact that the supplementary supervision introduced by the co-teaching models may inherently propagate errors and misclassify certain pseudo-labels. 
Nevertheless, setting the forget rate to 40\% yields the highest average performance across all 15 datasets, demonstrating that effectively reducing pseudo-label noise can further boost detection efficacy. 
These findings suggest that developing more robust and reliable pseudo-label refinement strategies constitutes a promising avenue for future research.

\subsection{Ablation Study.} 
\begin{table*}[htbp]
\caption{The evaluation results of ablation experiments across the datasets.}
\begin{adjustbox}{width=\textwidth}
\begin{tabular}{c|ccc|ccc|ccc|ccc|ccc}
\hline
\multirow{2}{*}{} & \multicolumn{3}{c|}{\textit{w/o aux}} & \multicolumn{3}{c|}{\textit{w/o contra}} & \multicolumn{3}{c|}{\textit{w/o adapt}} & \multicolumn{3}{c|}{\textit{w/o TTCL}} & \multicolumn{3}{c}{RTTAD}\\ 
\cline{2-16} 
 & auc-roc & auc-pr & f1
 & auc-roc & auc-pr & f1
 & auc-roc & auc-pr & f1
 & auc-roc & auc-pr & f1
 & auc-roc & auc-pr & f1 \\
 \hline
arrhythmia &
0.5808 & 0.5295 & 0.4394 & 
0.6195 & 0.5795 & 0.5152 & 
0.4509 & 0.4032 & 0.303 & 
0.7437 & 0.5612 & 0.5 & 	
0.819 & 0.6212 & 0.5455	 \\

breastw &
0.9833 & 0.9882 & 0.9414 & 
0.9856 & 0.9897 & 0.9582 & 
0.9841 & 0.9886	& 0.9498 &
0.99 & 0.9882 & 0.9498 & 	
0.9933 & 0.9921 & 0.9582 \\

cardio &
0.7744 & 0.6366 & 0.5966 & 
0.7283 & 0.6246 & 0.5966 & 
0.7309 & 0.6382 & 0.5852 & 
0.7106 & 0.2561 & 0.2273 & 	
0.8199 & 0.6385 & 0.6023 \\

cardiotocography &
0.3707 & 0.3575 & 0.279 & 
0.3122 & 0.3368 & 0.2725 & 
0.2981 & 0.3231 & 0.2618 & 
0.5941 & 0.3853 & 0.3691 & 	
0.6402 & 0.3906 & 0.4099 \\

glass &
0.2726 & 0.0858 & 0.1111 & 
0.083 & 0.0649 & 0. & 
0.2756 & 0.0797 & 0. & 	
0.6974 & 0.1596 & 0.2222 & 	
0.7021 & 0.1504 & 0.1111 \\

ionosphere &
0.5012 & 0.5321 & 0.627 & 
0.7009 & 0.7124 & 0.6905 & 
0.4966 & 0.53 & 0.6429 & 
0.715 & 0.7432 & 0.6349 & 	
 0.7446 & 0.7552 & 0.6508 \\

mammography &
0.7249 & 0.1763 & 0.2577 & 
0.7954 & 0.1955 & 0.2769 & 
0.8856 & 0.5099 & 0.5692 & 
0.8658 & 0.5074 & 0.5192 & 	
0.897 & 0.5407 & 0.5462	\\

optdigits &
0.2723 & 0.037 & 0. & 
0.3683 & 0.0425 & 0. & 
0.6349 & 0.0705 & 0. & 
0.8171 & 0.1337 & 0.0733 & 	
0.946 & 0.4199 & 0.4733	\\

pendigits &
0.9471 & 0.7146 & 0.6731 & 
0.7514 & 0.0814 & 0. & 
0.8837 & 0.2974 & 0.3526 & 	
0.4142 & 0.0378 & 0.0192 & 	
0.7313 & 0.0964 & 0.1154 \\

pima &
0.4721 & 0.5484 & 0.4851 & 
0.478 & 0.55 & 0.4813 & 
0.4704 & 0.5493 & 0.4851 & 	
0.592 & 0.6311 & 0.6045 & 
0.5724 & 0.5921 & 0.5858 \\

satellite &
0.6171 & 0.7131 & 0.5319 & 
0.4813 & 0.7486 & 0.5648 & 
0.5581 & 0.6679 & 0.4641 & 	
0.8054 & 0.8409 & 0.7279 & 	
0.8143 & 0.8307 & 0.721	
 \\

satimage-2 &
0.9876 & 0.945 & 0.9014 & 
0.9855 & 0.9383 & 0.9014 & 
0.9934 & 0.9385 & 0.8794 & 
0.9983 & 0.9718 & 0.9296 & 	
0.9985 & 0.9716 & 0.9296	
 \\

thyroid &
0.7541 & 0.3503 & 0.3333 & 
0.4189 & 0.0638 & 0.043 & 
0.2801 & 0.0349 & 0. & 
0.864 & 0.3749 & 0.3763 & 	
0.963 & 0.7773 & 0.6667 \\

wbc &
0.9448 & 0.718 & 0.7143 & 
0.9448 & 0.718 & 0.7143 & 
0.9448 & 0.718 & 0.7143 & 
0.9356 & 0.6493 & 0.6667 & 	
0.9723 & 0.8147 & 0.7143 \\

wine &
0.9983 & 0.9909 & 0.9 & 
0.9983 & 0.9909 & 0.9 & 
0.9983 & 0.9909 & 0.9 & 
0.9986 & 0.9909 & 0.9 & 	
0.9986 & 0.9909 & 0.9 \\
\hline
Average value &
0.6800 & 0.5548 & 0.5194 & 
0.6434 & 0.5091 & 0.4609 & 
0.6590 & 0.5160 & 0.4738 & 
0.7827 & 0.5487 & 0.5146 & 
\textbf{0.8408} & \textbf{0.6388} & \textbf{0.5953} \\

\hline
\end{tabular}
\end{adjustbox}
\label{tab:ablation_results}
% \vspace{-1.5em}
\end{table*}
To rigorously validate the efficacy of each core component within RTTAD, we design four variants for our ablation study:
\begin{itemize}
    \item \textbf{\textit{w/o aux}}: Removes the auxiliary feature reconstruction task during both the training and testing phases, aiming to evaluate the contribution of multi-level feature capture to detection performance.
    \item \textbf{\textit{w/o TTCL}}: Removes the entire test-time contrastive learning module, thereby regressing the framework to a static detector, to verify whether test-time distribution adaptation effectively mitigates the performance degradation inherent in static models.
    \item \textbf{\textit{w/o adapt}}: Adapts the model solely to the current test samples during the testing phase without retaining the previously learned normal patterns, exploring whether the absence of explicit constraints triggers catastrophic forgetting.
    \item \textbf{\textit{w/o contra}}: Eliminates the KNN-based contrastive optimization within the embedding space during test-time adaptation, investigating whether the pseudo-label-guided representation optimization genuinely bolsters the model's capability to discriminate between normal and anomalous samples.
\end{itemize}

The detailed quantitative results of these ablation experiments are tabulated in~\cref{tab:ablation_results}.
The full RTTAD framework consistently achieves the best performance across all evaluation metrics. 
Through an in-depth comparative analysis, we draw the following key conclusions.
First, the performance drop observed in \textbf{\textit{w/o aux}} indicates that the multi-level features provided by the auxiliary task are crucial for establishing a robust initial representation of normal patterns, serving as the solid foundation for subsequent test-time adaptation. 
Second, the performance of \textbf{\textit{w/o TTCL}} corroborates that when confronted with distribution shifts, static models struggle to identify anomalies effectively and inevitably suffer from performance degradation, whereas test-time adaptation can restore and enhance the model's discriminative capability. 
Most importantly, the detection performance of both \textbf{\textit{w/o adapt}} and \textbf{\textit{w/o contra}} is substantially inferior even to that of \textbf{\textit{w/o TTCL}}, which performs no adaptation at all. 
This phenomenon underscores the paramount importance of accounting for the presence of latent anomalies and the forgetting of prior knowledge during the testing phase. 
Blind adaptation devoid of risk awareness will severely compromise the model's intrinsic detection capability.

\subsection{Parameter Sensitivity Analysis.}
\begin{figure}[htbp]
% \vspace{-0.9em}
\begin{center}
\includegraphics[width=\columnwidth]{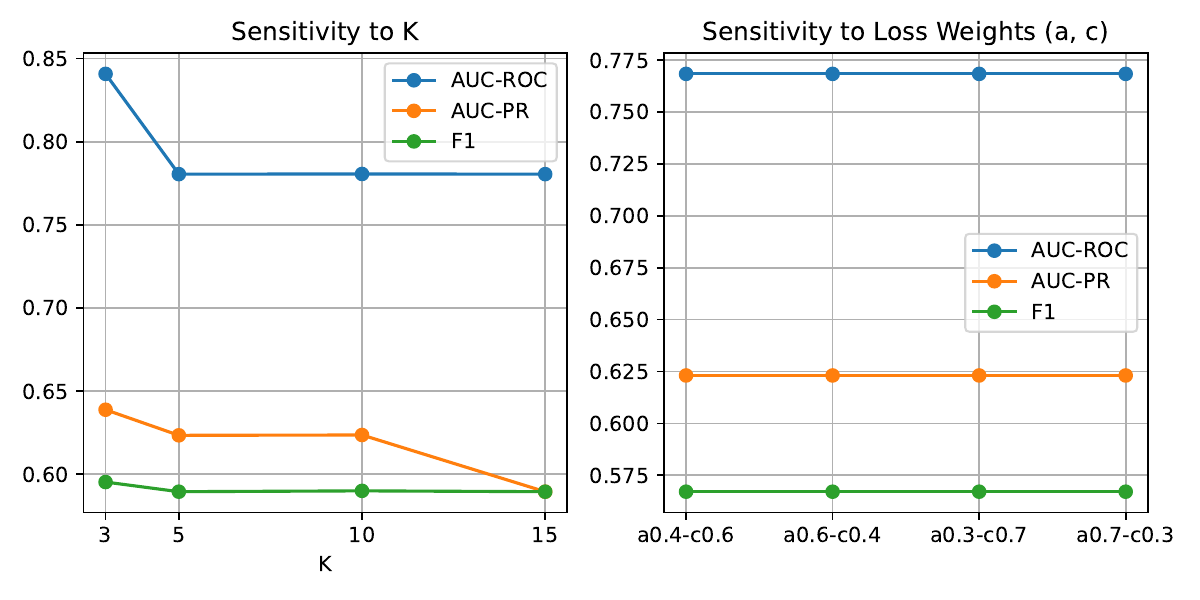}
\end{center}
 % \vspace{-0.5em}
\caption{Average detection performance across 15 datasets under different parameter settings}
 \label{fig:sensitivity}
 % \vspace{-1em}
\end{figure}
We conducted a parameter sensitivity analysis with respect to two key factors: the value of K used in the KNN-based contrastive learning module, and the weighting coefficients of the adaptation loss and contrastive loss. The resulting performance trends are presented in~\cref{fig:sensitivity}.

\textbf{Sensitivity to Neighborhood Size $K$.} The model achieves peak performance across all metrics at $K=3$, maintains high stability within $K \in [5, 10]$, and exhibits a slight decline at $K=15$. This trend is intuitive: a smaller $K$ precisely captures the highly localized manifold structure of normal samples, avoiding the erroneous forced alignment of distinct normal clusters. Conversely, given the multi-pattern nature of normal data, an excessively large $K$ inadvertently incorporates cross-cluster noise, which blurs representation boundaries and compromises the model's discriminative capability. Consequently, $K=3$ is adopted as the default configuration to strike the optimal balance.

\textbf{Sensitivity to Loss Weights.} Let $a$ denote the weight of $\mathcal{L}_{adapt}$ and $c$ denote the weight of $\mathcal{L}_{contra}$. The AUC-ROC, AUC-PR, and F1 scores exhibit consistency across all configurations, yielding nearly flat trajectories. This profound indicates a synergy between the adaptation and contrastive objectives, corroborating that RTTAD's superiority is driven by its core architectural design rather than meticulous hyperparameter tuning. Such insensitivity to weight variations alleviates the tuning burden in practical unsupervised scenarios devoid of validation data, demonstrating the framework's robustness and utility.

\section{Conclusion}
In this paper, we investigate the severe performance degradation of unsupervised tabular anomaly detection when confronting normality shifts at test time. Existing static methods suffer from an incomplete characterization of normality, rendering them highly vulnerable when normal patterns evolve. Furthermore, we demonstrate that directly applying indiscriminate test-time adaptation inevitably triggers catastrophic anomaly contamination due to the lack of reliable supervision and the inherent presence of latent anomalies. To address this dilemma, we propose RTTAD, a risk-aware test-time adaptation framework that holistically manages distribution shifts through a synergistic two-stage paradigm. During the training phase, RTTAD utilizes collaborative dual-task learning to capture multi-level representations, establishing a robust anchor of normal patterns. During the testing phase, it executes a rigorous risk-controlled adaptation via test-time contrastive learning. This mechanism safely adapts to shifted normal distributions while explicitly suppressing anomaly contamination to bolster the model's intrinsic discriminative capability. Extensive experiments on both synthetic and temporally evolving normality shifts demonstrate that RTTAD consistently achieves state-of-the-art robust detection performance. Future work will explore more reliable pseudo-label refinement and dynamic risk estimation strategies to further enhance the safety of test-time adaptation in open-world scenarios.

\section*{Acknowledgments}
This work was supported by the National Natural Science Foundation of China (No.72274022 \& No.6240073908), the Beijing Natural Science Foundation (No.4244083) and the Fundamental Research Funds for the Central Universities (No.500422828).

\bibliographystyle{IEEEtran}
\bibliography{reference}

\vfill

\end{document}